\documentclass[conference, a4paper]{IEEEtran}
\IEEEoverridecommandlockouts
\usepackage{cite}
\usepackage{amsmath,amssymb,amsfonts}
\usepackage{algorithmic}
\usepackage{graphicx}
\usepackage{textcomp}
\usepackage{caption}
\usepackage{subcaption}
\usepackage{xcolor, makecell,multirow,layouts,epsfig}
\def\BibTeX{{\rm B\kern-.05em{\sc i\kern-.025em b}\kern-.08em
    T\kern-.1667em\lower.7ex\hbox{E}\kern-.125emX}}

\begin{document}

\title{Plug n' Play Channel Shuffle Module for Enhancing Tiny Vision Transformers}
\author{\IEEEauthorblockN{1\textsuperscript{st} Xuwei Xu}
\IEEEauthorblockA{\textit{School of Electrical Engineering and Computer Science} \\
\textit{The University of Queensland}\\
Brisbane, Australia\\
xuwei.xu@uq.edu.au}
\and
\IEEEauthorblockN{2\textsuperscript{nd} Sen Wang}
\IEEEauthorblockA{\textit{School of Electrical Engineering and Computer Science} \\
\textit{The University of Queensland}\\
Brisbane, Australia\\
sen.wang@uq.edu.au}
\and
\IEEEauthorblockN{3\textsuperscript{rd} Yudong Chen}
\IEEEauthorblockA{\textit{School of Electrical Engineering and Computer Science} \\
\textit{The University of Queensland}\\
Brisbane, Australia\\
yudong.chen@uq.edu.au}
\and
\IEEEauthorblockN{4\textsuperscript{th} Jiajun Liu}
\IEEEauthorblockA{\textit{DATA61} \\
\textit{Commonwealth Scientific and Industrial Research Organisation}\\
Brisbane, Australia\\
ryan.liu@data61.csiro.au}
}

\maketitle

\begin{abstract}
Vision Transformers (ViTs) have demonstrated remarkable performance in various computer vision tasks. However, the high computational complexity hinders ViTs' applicability on devices with limited memory and computing resources. Although certain investigations have delved into the fusion of convolutional layers with self-attention mechanisms to enhance the efficiency of ViTs, there remains a knowledge gap in constructing tiny yet effective ViTs solely based on the self-attention mechanism. Furthermore, the straightforward strategy of reducing the feature channels in a large but outperforming ViT often results in significant performance degradation despite improved efficiency. To address these challenges, we propose a novel channel shuffle module to improve tiny-size ViTs, showing the potential of pure self-attention models in environments with constrained computing resources. Inspired by the channel shuffle design in ShuffleNetV2 \cite{ma2018shufflenet}, our module expands the feature channels of a tiny ViT and partitions the channels into two groups: the \textit{Attended} and \textit{Idle} groups. Self-attention computations are exclusively employed on the designated \textit{Attended} group, followed by a channel shuffle operation that facilitates information exchange between the two groups. By incorporating our module into a tiny ViT, we can achieve superior performance while maintaining a comparable computational complexity to the vanilla model. Specifically, our proposed channel shuffle module consistently improves the top-1 accuracy on the ImageNet-1K dataset for various tiny ViT models by up to 2.8\%, with the changes in model complexity being less than 0.03 GMACs.

\end{abstract}


\section{Introduction}
Vision Transformers (ViTs) have dominated the computer vision area since the success of \cite{dosovitskiy2020image}, demonstrating remarkable performance in image classification \cite{dosovitskiy2020image, zhai2022scaling, ding2022davit}, object detection \cite{liu2021swin, liu2022swin, li2022exploring} and segmentation \cite{yang2021focal, chen2022vision, fang2023eva}. However, the high computational burden of the self-attention mechanism makes ViTs less efficient compared to traditional convolutional neural networks (CNNs) \cite{howard2017mobilenets, sandler2018mobilenetv2, ma2018shufflenet, zhang2018shufflenet, tan2019efficientnet} on devices with constrained memory and computing resources. As a result, there is a growing interest in the research community to develop lightweight and efficient ViT models.

Various approaches have been proposed to address this challenge. Some methods integrate efficient convolution operations with computationally expensive self-attentions to create hybrid efficient ViTs \cite{chen2021visformer, wu2021cvt, dai2021coatnet,mehta2021mobilevit, graham2021levit, d2021convit}. However, these methods do not fully exploit the potential of pure self-attention models to achieve both high performance and efficiency. Alternatively, certain studies revisit the design principles of efficient CNNs and transfer them to the design of efficient ViTs, such as window-based attention \cite{liu2021swin, dong2021cswin}, hierarchical network architecture \cite{wang2021pyramid, liu2021swin}, bottleneck structure \cite{srinivas2021bottleneck} and spatially separable self-attention \cite{chu2021twins}. It is worth noting that the channel shuffle design introduced by \cite{zhang2018shufflenet} is less explored in this context. In addition, some powerful ViT models construct their lightweight versions by simply reducing the number of feature channels, layers, or self-attention heads \cite{touvron2021training,yuan2021tokens,liu2021swin,wang2021pyramid}. However, as demonstrated in Table \ref{tab:lightweight comparison}, such a naive model size reduction often leads to a significant performance drop. For example, DeiT-Tiny \cite{touvron2021training} suffers a 7.7\% top-1 accuracy drop on the ImageNet \cite{deng2009imagenet} compared to DeiT-Small when the number of feature channels declines from 384 to 192. 

\begin{table}[t]
\caption{Comparisons of pure ViT models on ImageNet \cite{deng2009imagenet} validation set. * indicates that the model is re-trained on our machine. The number of feature channels of Swin Transformer \cite{liu2021swin} only indicates the first stage's feature channels, which would expand as the layer goes deep}
\label{tab:lightweight comparison}
\setlength{\tabcolsep}{3pt}
\begin{center}
\begin{tabular}{|l|c|c|c|c|c|}
    \hline
    Methods & \makecell{Feature\\Channels} & Layers & Param. & MACs & \makecell{Top-1\\Acc.} \\
    \hline
    T2T-ViT-7 \cite{yuan2021tokens} & 256 & 7 & 4.3M & 1.1G & 71.7\% \\
    T2T-ViT-14 \cite{yuan2021tokens} & 384 & 14 & 21.5M & 4.8G & 81.7\% \\
    \hline
    DeiT-Tiny \cite{touvron2021training} & 192 & 12 & 5.7M & 1.3G & 72.2\% \\
    DeiT-Small \cite{touvron2021training} & 384 & 12 & 21.8M & 4.6G & 79.9\% \\
    \hline
    Swin-ExtraTiny* \cite{liu2021swin} & 48 & 12 & 6.8M & 1.1G & 74.8\% \\
    Swin-Tiny \cite{liu2021swin} & 96 & 12 & 28.8M & 4.5G & 80.8\% \\
    \hline  
\end{tabular}
\end{center}
\end{table}
We figure out that one of the main reasons for the performance degradation in tiny ViTs is the limited number of feature channels. The insufficient number of feature channels makes tiny ViTs unable to represent the image effectively. To mitigate the similar issue of insufficient image representations, previous studies in efficient CNNs leverage the concept of grouped convolution \cite{krizhevsky2012imagenet, xie2017aggregated, chollet2017xception}, which reduces computational complexity and memory footprint without compromising the total number of feature channels. Besides, \cite{zhang2018shufflenet} proposes a channel shuffle operation to help the information flow across groups. And \cite{ma2018shufflenet} extensively explores the architecture design and introduces a strategy that splits the channels into two groups, allowing one group to remain idle throughout the layer and shuffling channels between the two groups. It is worth noting that these designs in efficient CNNs are seldom introduced to efficient pure self-attention models.

Hence, in this paper, we present a channel shuffle module specifically designed for tiny ViT models to address the aforementioned challenges. Inspired by \cite{zhang2018shufflenet, ma2018shufflenet}, our module expands the feature channels of a compressed ViT model and separates them into two groups, namely the \textit{Attended} group and the \textit{Idle} group. In each layer, the \textit{Attended} group performs self-attention computation like a conventional ViT while the \textit{Idle} group remains inactive during the computation. At the end of each layer, a channel shuffle operation is employed to interleave the two feature channel groups and facilitate information exchange. This module serves as a plug-and-play enhancement to tiny ViTs, which improves the performance with merely a bit more computations. Meanwhile, our module is generic and can be applied to both plain and hierarchical ViTs. In this paper, we select DeiT \cite{touvron2021training} and T2T-ViT \cite{yuan2021tokens} as representatives for plain ViTs, and Swin Transformer \cite{liu2021swin} as the representative for hierarchical ViTs. Moreover, we observe that the \textit{Idle} channels may exhibit different scales compared to the \textit{Attended} channels, resulting in many trivial channels after Layer Normalization. To address this issue, we propose a simple channel re-scaling optimization to alleviate the problem. Extensive experiments have demonstrated the efficacy and efficiency of our module.

We summarize the key contributions of our work as follows:
\begin{itemize}
    \item We develop an efficient channel shuffle module to enhance tiny ViTs with very few additional computations, satisfying the environment with constrained computing resources.
    \item Our module can work as an independent plug-and-play component to the vanilla tiny ViTs and is generic for both plain and hierarchical ViTs.
    \item We introduce a simple channel re-scaling method to mitigate the problem of distinct scales between the \textit{Attended} and \textit{Idle} groups.
    \item Extensive experimental results have shown the efficiency and efficacy of our proposed module.
\end{itemize}
To our best knowledge, this is the first work improving tiny-scale efficient ViTs by enriching channel-wise information while maintaining computational complexity.

\section{Related work}
\subsection{Efficient convolutional neural networks }
Efficient convolutional neural networks (CNNs) have attracted significant attention due to the need for deployment on devices with constrained computing resources. AlexNet \cite{krizhevsky2012imagenet} proposes grouped convolution to distribute the model over multiple GPUs and consequently reduce the computational complexity and memory footprint on each single GPU. ResNeXt \cite{xie2017aggregated} validates the efficacy of grouped convolution, showing an improved accuracy by grouped convolution. GoogLeNet \cite{szegedy2015going} leverages grouped convolution to establish the inception module, which successfully expands the width and depth of a CNN model while keeping the computational budget constant. Xception \cite{chollet2017xception} enforces the number of channel groups the same as the number of feature channels and brings up the concept of depth-wise separable convolution. MobileNets \cite{howard2017mobilenets, sandler2018mobilenetv2} extensively utilize depth-wise separable convolution to reduce the computational complexity of CNNs for mobile vision applications. ShuffleNet \cite{zhang2018shufflenet} puts forth the concept of channel shuffle for grouped convolution, which realizes efficient information exchange between channel groups by shuffling the channels. These efficiency designs are less explored in ViTs than in CNNs.

\subsection{Vision Transformers}
Vision Transformers (ViTs) have gained significant attention in the field of computer vision as a promising alternative to CNNs. The original ViT architecture \cite{dosovitskiy2020image} demonstrates the effectiveness of the self-attention mechanism for image classification, which is capable of capturing global relationships. In general, there are two types of ViT architectures, namely plain and hierarchical, which are distinguished by whether token downsampling is adopted in the network. Plain ViTs \cite{touvron2021training, yuan2021tokens, zhai2022scaling} have the same backbone architecture as the vanilla ViT that the number of tokens and feature channels keeps static throughout the network. Hierarchical ViTs \cite{wang2021pyramid, liu2021swin, dong2021cswin, liu2022swin} apply token downsampling between stages to enable multi-scale self-attention. As a result, the number of tokens decreases and the number of feature channels increases as the layer goes deep in hierarchical ViTs. In this work, we choose DeiT \cite{touvron2021training} and T2T-ViT \cite{yuan2021tokens} as representatives for plain ViTs, and Swin Transformer \cite{liu2021swin} as the representative for hierarchical ViTs.

\subsection{Efficient Vision Transformers}
One direction of ViT research focuses on improving the efficiency of ViTs, as their computational complexity can be prohibitive for resource-constrained devices. Several approaches have been explored, such as distillation methods that transfer knowledge from large ViT models to smaller ones \cite{touvron2021training}, spatial-wise token pruning \cite{rao2021dynamicvit, liang2021evit, xu2022evo}, and regional self-attention design \cite{liu2021swin, dong2021cswin}. These methods mainly concentrate on compressing a powerful ViT into a smaller counterpart without compromising much performance, while our method attempts to enhance the tiny ViTs. Besides, some studies \cite{graham2021levit, dai2021coatnet, wu2021cvt, chen2021visformer, mehta2021mobilevit, d2021convit} integrate convolution with self-attention to achieve efficient ViTs. However, these methods fail to reveal the potential of pure self-attention models in an environment with limited computing resources. In contrast, our module demonstrates the possibility of efficient and high-performance pure self-attention models.

\section{Methods}
\begin{figure*}
    \centering
    \begin{subfigure}[b]{0.8\linewidth}
        \centering
        \includegraphics[width=\linewidth]{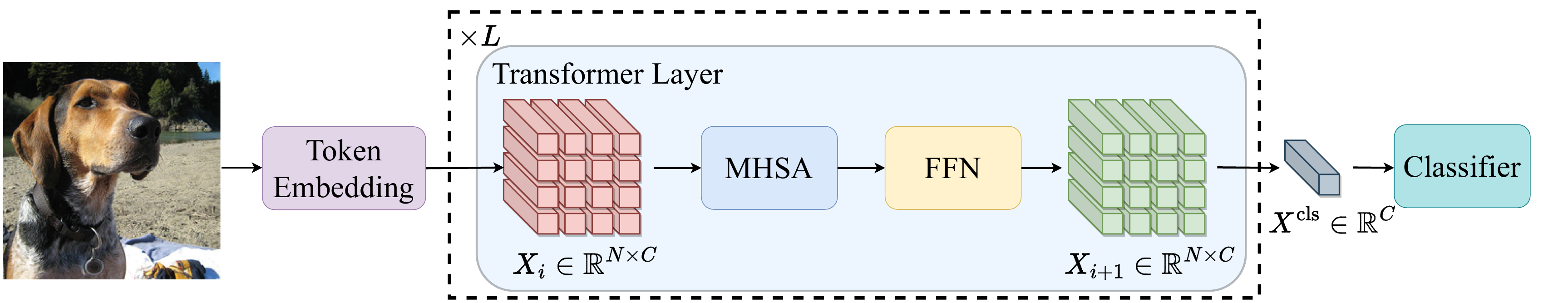}
        \caption{Vanilla Vision Transformer.}
        \vspace{1em}
    \end{subfigure}
    \begin{subfigure}[b]{\linewidth}
        \centering
        \includegraphics[trim={0 0 17em 0}, clip, width=\linewidth]{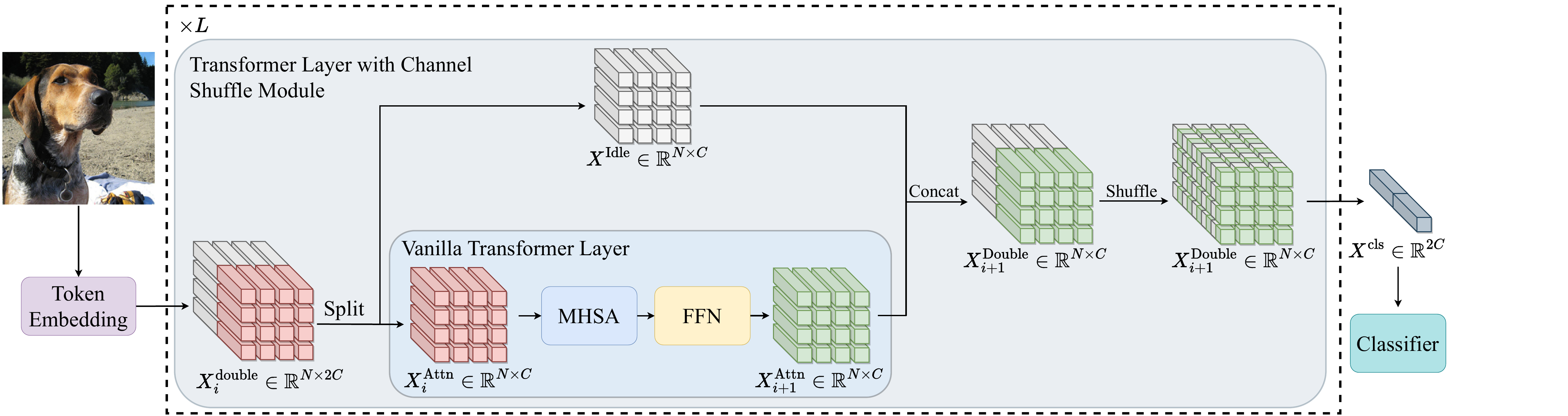}
        \caption{Vision Transformer with channel shuffle module.}
     \end{subfigure}
    \caption{Overview of the channel shuffle module. In each ViT layer, the \textit{Attended} group participates in the computations while the \textit{Idle} group retains the same until the end of the layer. At the end of each layer, we concatenate the two groups and shuffle the channels to facilitate information exchange.}
    \label{fig:shufflemodule}
\end{figure*}
\subsection{Preliminaries}
The vanilla ViT \cite{dosovitskiy2020image} first splits the input image into patches and then linearly projects the image patches into image tokens. These tokens serve as the input for subsequent computations, enabling the model to capture global contextual information. We denote the input feature map of layer $i$ as $X_i\in \mathbb{R}^{N \times C}$, where $N$ and $C$ are the numbers of tokens and feature channels, respectively. Each ViT layer comprises a multi-head self-attention (MHSA) module and a feed-forward network (FFN) module. For the MHSA module, it linearly transforms the input feature map into three matrices called \textit{Key} ($K_i$), \textit{Query} ($Q_i$) and \textit{Value} ($V_i$) by
 \begin{equation}
    K_i = X_iW_{k_{i}}, \quad Q_i = X_iW_{q_{i}}, \quad V_i = X_iW_{v_{i}},
\end{equation}
where $W_{k_{i}}$, $W_{q_{i}}$ and $W_{v_{i}}$ are the corresponding weights and the bias terms are omitted. Next, it computes the attention map $A_i\in \mathbb{R}^{N \times N}$ by a dot production with the softmax activation between \textit{Key} and \textit{Query} as
\begin{equation}
    A_{i}=\text{softmax}(\frac{Q_iK_i^\top}{\sqrt{d_k}}),
\end{equation}
where $d_k$ is the dimension of $K$ and is usually the same as the channel dimension $C$. The attention map reflects the similarities between each pair of tokens. Eventually, MHSA calculates the attended output with residual connection as
\begin{equation}
    \label{eq:mhsa}
    X_{i}'=\text{MHSA}(X_{i})+X_{i}=A_{i}V_{i}W_{m}+X_{i},
\end{equation}
where $W_{m}\in \mathbb{R}^{C \times C}$ is the learnable weight. After the MHSA module, a two-layer Multi-Layer Perceptron (MLP) is utilized as the FFN module to self-activate each token by
\begin{equation}
    \label{eq:ffn}
    X_{i+1}= \text{FFN}(X_{i}’)+X_{i}’=X_{i}’W_{f1}W_{f2}+X_{i}’,
\end{equation}
where $W_{f1}\in \mathbb{R}^{C \times \mu C}$ and $W_{f2}\in \mathbb{R}^{\mu C \times C}$ are two learnable projection weights, $\mu$ is the channel expansion ratio, $X_{i+1} \in \mathbb{R}^{N \times C}$ is the output of the $i^{\text{th}}$ ViT layer and the bias terms are omitted. Moreover, ViT introduces pre-module Layer Normalization (LN) \cite{ba2016layer} on the feature map to improve both the training time and the generalization performance.

\subsection{Channel shuffle module}
Fig. \ref{fig:shufflemodule} provides an overview of our proposed channel shuffle module, which aims to enhance the capabilities of tiny ViT models. To improve the feature representation ability, we start by doubling the feature channels of the input feature map $X_i\in \mathbb{R}^{N \times C}$ during the token embedding phase, resulting in $X_i^{\text{Double}}\in \mathbb{R}^{N \times 2C}$. In the $i^{\text{th}}$ layer, the feature map $X_i^{\text{Double}}$ is partitioned into two groups alone the feature channel dimension, namely the \textit{Attended} group $X_i^{\text{Attn}}\in \mathbb{R}^{N \times C}$ and \textit{Idle} group $X_i^{\text{Idle}}\in \mathbb{R}^{N \times C}$. The \textit{Attended} group only occupies half of the channels and participates in this layer's calculations, resulting in the attended output $X_{i+1}^{\text{Attn}}\in \mathbb{R}^{N \times C}$. On the contrary, the \textit{Idle} group holds the rest half channels and maintains the same until the end of this layer so that $X_i^{\text{Idle}}=X_{i+1}^{\text{Idle}}\in \mathbb{R}^{N \times C}$. At the end of layer $i$, the two groups are concatenated together as
\begin{equation}
    X_{i+1}^{\text{Double}}=\text{concat}(X_{i+1}^{\text{Attn}}, X_{i+1}^{\text{Idle}}),
\end{equation}
where the $X_{i+1}^{\text{Double}}\in \mathbb{R}^{N \times 2C}$ is the output of layer $i$. Finally, we apply channel shuffle on $X_{i+1}^{\text{Double}}$ to enforce information exchange between the two groups.

\subsection{Channel re-scaling}
We have identified a potential issue related to the residual connections \cite{he2016deep} employed in the MHSA and FFN modules, which can result in significant scale differences between $X_{i+1}^{\text{Attn}}$ and $X_{i+1}^{\text{Idle}}$. This discrepancy can lead to many trivial values after Layer Normalization, particularly in deeper layers. To address this issue, we devise a channel re-scaling approach for $X_{i+1}^{\text{Attn}}$. Specifically, we modify the Transformer layer described in Equations \ref{eq:mhsa} and \ref{eq:ffn} as follows:
\begin{equation}
    \begin{aligned}
    X_{i}^{\text{Attn}'}=&\text{MHSA}(X_{i}^{\text{Attn}})+\alpha_1 X_{i}^{\text{Attn}},\\
    X_{i+1}^{\text{Attn}}=&\text{FFN}(X_{i}^{\text{Attn}'})+\alpha_2 X_{i}^{\text{Attn}'},
    \end{aligned}
\end{equation}
where $\alpha_1, \alpha_2 \in \mathbb{R}^{C}$ are two learnable coefficients. We can use this simple modification to enable our module automatically control the scale of the feature representation in the \textit{Attended} group.

\subsection{Computational complexity analysis}
We begin by comparing the theoretical computational complexity of a vanilla ViT and a tiny ViT equipped with our proposed module, assuming both models have an equal number of total feature channels. In the case of the vanilla ViT, the computational complexity per layer can be represented as
\begin{equation}
    \Omega(\text{vanilla}) = (4+2\mu)NC^2+2N^2C,\\
\end{equation}
where $N, C$ and $\mu$ denote the number of tokens, feature channels and the expansion ratio of the MLP, respectively. On the other hand, the computational complexity of a tiny ViT with our channel shuffle module is given by 
\begin{equation}
    \Omega(\text{shuffle}) = (1+\frac{\mu}{2})NC^2+N^2C+NC.
\end{equation}
Notably, the computational complexity of our shuffle module is significantly smaller than that of the vanilla ViT. Therefore, when considering models with an equal number of total feature channels, indicating a similar feature representation capacity, our proposed module proves to be more computationally efficient and suitable for deployment on resource-constrained devices.

Furthermore, we analyze the overall computational complexity of a tiny ViT model with and without the channel shuffle module. Since the \textit{Idle} group does not participate in computations, the increase in computational complexity per layer in our module solely stems from the channel re-scaling, which amounts to $NC$. Consequently, the total increase in computational complexity over $L$ layers is $LNC$. Additionally, our module doubles the channels during the token embedding phase, resulting in additional $3P^2NC$ computations, where $P$ represents the size of the image patches. Moreover, in the final layer, our module introduces an extra $SC$ computations, with $S$ denoting the number of classes. As a result, the total increase in computational complexity is given by $LNC + 3P^2NC + SC$, which is relatively small compared to the overall computational complexity of $3P^2NC+(4+2\mu)LNC^2+2LN^2C+SC$. Taking DeiT-Tiny \cite{touvron2021training} as an example, the total computational complexity is approximately 1.25GMACs while the extra computational complexity brought up by our channel shuffle module is merely 0.03GMACs, which is about 2\% of the total.

\subsection{Plug-and-play module}
An important contribution of this module is its ability to address the performance degradation of tiny ViT models. The channel shuffle module can be easily incorporated into a tiny ViT without introducing significant modifications to the backbone architecture. As depicted in Fig. \ref{fig:shufflemodule}, the \textit{Idle} group does not participate in calculations, making it straightforward to apply the module to different variants of ViT architectures.

When integrating the channel shuffle module into a plain ViT, the number of feature channels is doubled during the image token embedding phase and remains constant throughout the network. However, in hierarchical ViTs, the computational complexity can be proportional to the square of the number of feature channels in downsampling layers. Our channel shuffle module could lead to a non-negligible increase in computations. Taking Swin Transformer \cite{liu2021swin} as an instance, the computational complexity of a fully-connected downsampling layer is $2NC^2$. However, if the feature channels are doubled in the channel shuffle module, the computational complexity of downsampling would increase to $8NC^2$ per downsampling layer. To address this issue, we perform separate downsampling operations on the \textit{Attended} and \textit{Idle} groups to prevent an excessive computational burden when applying the channel shuffle module to hierarchical ViTs.

\section{Experiments}
\subsection{Dataset and settings}
\noindent \textbf{Dataset.}\quad We choose ImageNet-1K \cite{deng2009imagenet} as the target dataset, which contains around 1.28 million images for training and 50 thousand images for validation. It is an acknowledged standard dataset for model benchmarking.

\noindent \textbf{Base model.}\quad DeiT-Tiny \cite{touvron2021training} and T2T-ViT-7 \cite{yuan2021tokens} are selected as representatives for tiny ViTs with a plain architecture. Swin Transformer \cite{liu2021swin} is chosen to be the representative for hierarchical ViTs. Since Swin Transformer does not officially provide a mobile-level version whose computational complexity is approximately 1GMACs, we scale down the Swin-Tiny by reducing the number of its feature channels in the first stage from 96 to 48 and subsequently construct a Swin-ExtraTiny model. These are well-known in ViT families for their excellent standard performance and data-efficient training. Furthermore, we also reproduce the experiments on Swin-Ti to explore the influence of base model size.

\noindent \textbf{Training configurations.}\quad We follow the image augmentations and training recipes in \cite{touvron2021training} and its official GitHub repository for all the models, except setting the learning rate to 5e-3, the total batch size to 4096.

\begin{table}[t]
\caption{Effectiveness of the channel shuffle module. We compare the top-1 accuracy on ImageNet \cite{deng2009imagenet} of tiny ViT models equipped with the channel shuffle module and its vanilla version.}
\label{tab:mainresults}
\setlength{\tabcolsep}{2pt}
\begin{center}
\begin{tabular}{|l|c|c|c|c|c|}
    \hline
    Methods & \makecell{Feature\\Channels} & Layers & Param. & MACs & \makecell{Top-1\\Acc.} \\
    \hline
    T2T-ViT-7 \cite{yuan2021tokens} & 256 & 7 & 4.3M & 1.1G & 71.7\% \\
    Shuffled T2T-ViT-7 & 512 & 7 & 4.7M & 1.2G & 74.4\% (+2.7\%)\\
    \hline
    DeiT-Tiny \cite{touvron2021training} & 192 & 12 & 5.7M & 1.3G & 72.2\% \\
    Shuffled DeiT-Tiny & 384 & 12 & 6.1M & 1.3G & 74.4\% (+2.2\%) \\
    \hline
    Swin-ExtraTiny \cite{liu2021swin} & 48 & 12 & 6.8M & 1.1G & 74.8\% \\
    Shuffled Swin-ExtraTiny & 96 & 11 & 7.2M & 1.0G & 77.8\% (+3.0\%) \\
    \hline  
\end{tabular}
\end{center}
\end{table}
\subsection{Main results}
Table \ref{tab:mainresults} provides comparisons between pure self-attention models with and without our channel shuffle module in terms of accuracy, the number of parameters and computational complexity. The results clearly show that the channel shuffle module significantly enhances the performance of tiny ViTs in the classification task, improving the accuracy by 2.2$\sim$3.0\% without a significant increase in computational budgets. In the case of DeiT-Tiny, the shuffled version achieves a 2.2\% higher top-1 accuracy while maintaining an equivalent computational cost to the vanilla version. As for Swin-ExtraTiny, since we eliminate one layer, the shuffled version (1.0GMACs) runs faster than the unshuffled version (1.1GMACs) with a remarkable 3.0\% higher top-1 accuracy. These results highlight the effectiveness and generalizability of our simple yet powerful design module.

\begin{table}[t]
\caption{Comparisons against tiny-size CNNs and ViTs.}
\label{tab:comparisons}
\setlength{\tabcolsep}{3.5pt}
\begin{center}
\begin{tabular}{|l|l|c|c|c|}
    \hline
    Methods & Type & Param. & MACs & \makecell{Top-1\\Acc.} \\
    \hline
    MobileNet V2 \cite{sandler2018mobilenetv2} & CNN & 6.9M & 0.6G & 74.7\% \\
    ShuffleNet V2 \cite{ma2018shufflenet} & CNN & 7.4M & 0.6G & 74.9\% \\
    EfficientNet-B0 \cite{tan2019efficientnet} & CNN & 5.3M & 0.4G & 76.3\% \\
    EfficientNet-B1 \cite{tan2019efficientnet} & CNN & 7.8M & 0.7G & 79.1\% \\
    RegNetY-800MF \cite{radosavovic2020designing} & CNN & 6.3M & 0.8G & 76.3\% \\
    RegNetY-1.6GF \cite{radosavovic2020designing} & CNN & 11.2M & 1.6G & 78.0\% \\
    \hline
    T2T-ViT-7 \cite{yuan2021tokens} & ViT & 4.3M & 1.1G & 71.7\% \\
    HVT-Ti-1 \cite{pan2021scalable} & ViT & 5.6M & 0.7G & 69.6\% \\
    DeiT-Tiny \cite{touvron2021training} & ViT & 5.7M & 1.3G & 72.2\% \\
    PVTv2-B0 \cite{wang2022pvt} & ViT & 3.4M & 0.6G & 70.5\% \\
    PVTv2-B1 \cite{wang2022pvt} & ViT & 13.1M & 2.1G & 78.7\% \\
    PVTv1-Tiny \cite{wang2021pyramid} & ViT & 13.2M & 1.9G & 75.1\% \\
    AutoFormer-Tiny \cite{chen2021autoformer} & ViT & 5.7M & 1.3G & 74.7\% \\
    \hline 
    PiT-Ti \cite{heo2021rethinking} & Hybrid & 4.9M & 0.7G & 74.6\% \\
    ConViT-Ti \cite{d2021convit} & Hybrid & 6.0M & 1.0G & 73.1\% \\
    Visformer-Ti \cite{chen2021visformer} & Hybrid & 10.3M & 1.3G & 78.6\% \\
    MobileViTv1-XS \cite{mehta2021mobilevit} & Hybrid & 2.3M & 0.9G & 74.8\% \\
    MobileViTv1-S \cite{mehta2021mobilevit} & Hybrid & 5.6M & 2.0G & 78.4\% \\
    EdgeNeXt-XS \cite{maaz2022edgenext} & Hybrid & 2.3M & 1.1G & 75.0\% \\
    EdgeNeXt-S \cite{maaz2022edgenext} & Hybrid & 5.6M & 2.6G & 78.4\% \\
    \hline 
    Shuffled T2T-ViT-7 (ours) & ViT & 4.7M & 1.2G & 74.4\% \\
    Shuffled DeiT-Tiny (ours) & ViT & 6.1M & 1.3G & 74.4\% \\
    Shuffled Swin-ExtraTiny (ours) & ViT & 7.2M & 1.0G & 77.8\% \\
    \hline 
\end{tabular}
\end{center}
\end{table}
Table \ref{tab:comparisons} provides comprehensive comparisons between our channel shuffle module, other efficient CNNs, ViTs, and hybrid models. Firstly, our channel shuffle module consistently outperforms all other pure ViT models in terms of the trade-off between model complexity and accuracy, highlighting the effectiveness of our module. Secondly, when compared to hybrid models that combine convolution and self-attention, our module achieves comparable or even better performance. For instance, at the same model complexity of 1.1GMACs, Swin-ExtraTiny with the channel shuffle module outperforms EdgeNeXt-XS by 2.8\%. This result demonstrates the potential of efficient pure self-attention networks. However, it is worth noting that when compared to mobile-friendly CNNs, self-attention-based networks still exhibit lower efficiency when achieving the same performance.

\subsection{Feature channel analysis}
The channel shuffle module plays a critical role in enhancing the ability of a tiny ViT model to represent image features. Even though half of the channels do not actively participate in the Transformer layer, they still serve two important functions: propagating gradients across layers and enriching the available image features.

\begin{figure}[t]
\begin{minipage}[b]{.48\linewidth}
  \centering
  \centerline{\epsfig{figure=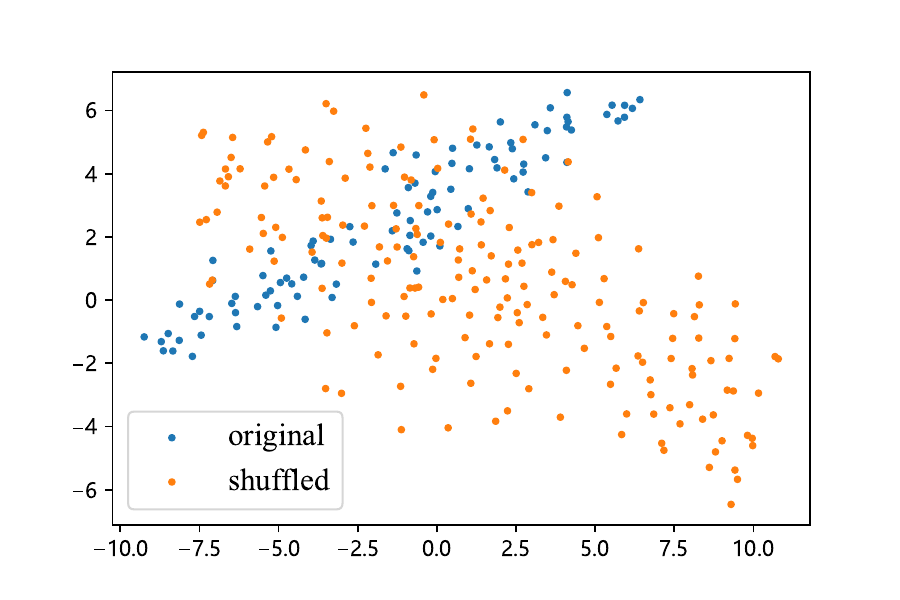,width=1.2\linewidth}}
  \centerline{\small (a) Stage 1}\medskip
\end{minipage}
\hfill
\begin{minipage}[b]{.48\linewidth}
  \centering
  \centerline{\epsfig{figure=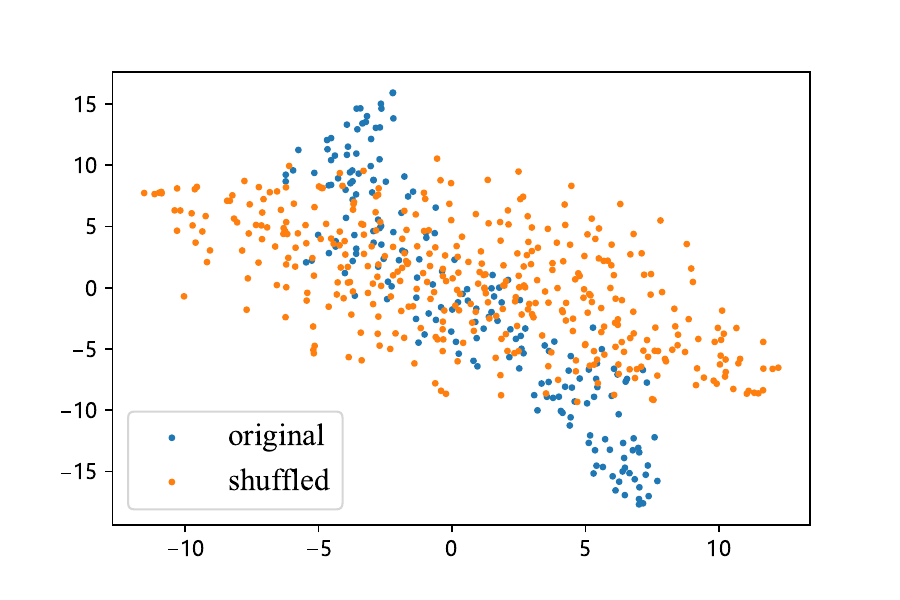,width=1.2\linewidth}}
  \centerline{\small (b) Stage 2}\medskip
\end{minipage}
\begin{minipage}[b]{.48\linewidth}
  \centering
  \centerline{\epsfig{figure=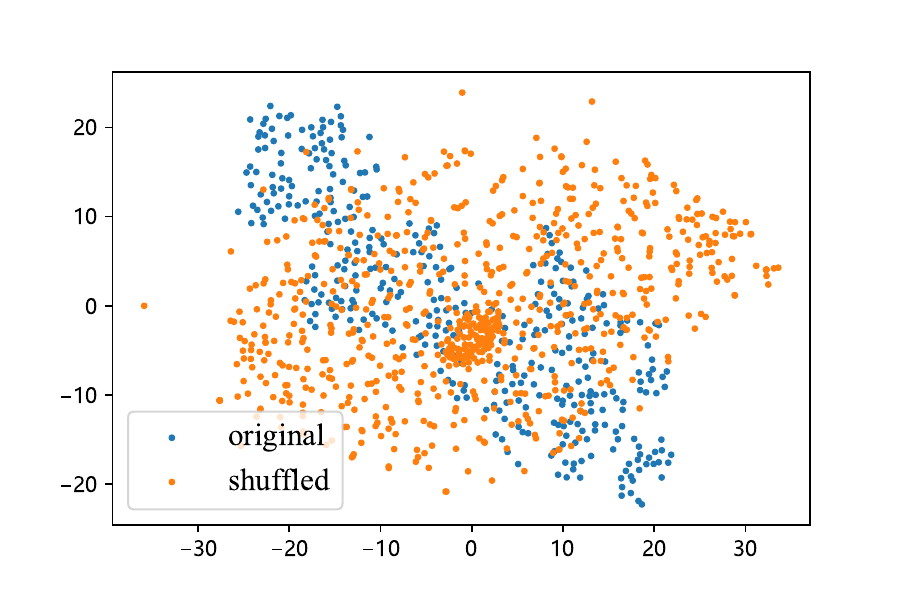,width=1.2\linewidth}}
  \centerline{\small (c) Stage 3}\medskip
\end{minipage}
\hfill
\begin{minipage}[b]{0.48\linewidth}
  \centering
  \centerline{\epsfig{figure=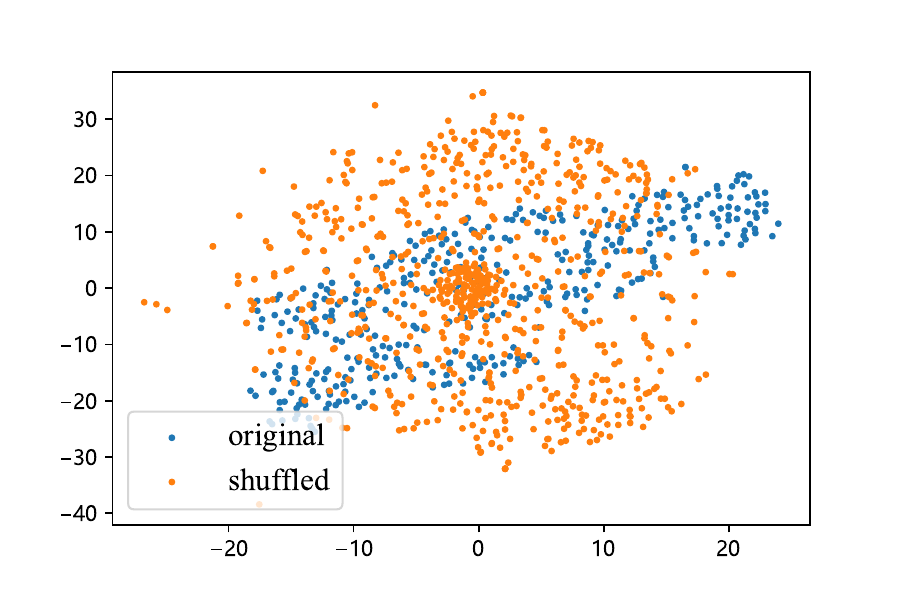,width=1.2\linewidth}}
  \centerline{\small (d) Stage 4}\medskip
\end{minipage}
\caption{Channel distributions of feature maps at the end of the four stages in Swin-ExtraTiny. The blue dots represent the distribution of the vanilla Swin-ExtraTiny while the orange dots stand for the shuffled version. Since Swin-ExtraTiny is a hierarchical vision Transformer, the number of channels (i.e., the number of dots in this figure) increases in deeper stages.}
\label{fig:distribution}
\end{figure}

To validate this argument, we visualize and compare the feature channel distributions with respect to the four stages in the shuffled and vanilla Swin-ExtraTiny in Fig. \ref{fig:distribution}. We employ t-distributed stochastic neighbour embedding (t-SNE) \cite{hinton2002stochastic} for dimensionality reduction. In Fig. \ref{fig:distribution}(a), it is evident that in the early stage with a small number of channels (e.g., 48 for Swin-ExtraTiny and 96 for shuffled Swin-ExtraTiny), the shuffled model (orange) exhibits a more diverse distribution compared to the unshuffled model (blue). This indicates that the shuffled model benefits from our module by incorporating richer information in the early stage. However, as the model progresses to deeper stages and the number of feature channels increases, the distribution variances between the shuffled and unshuffled models become less distinct, as shown in Fig. \ref{fig:distribution}(c) and \ref{fig:distribution}(d). We think this is the main reason why our channel shuffle module can improve the performance of tiny ViT models.

However, we also point out that the improvement of this module diminishes as the model size increases. Larger ViT models already possess sufficient feature channels to effectively represent the image, and excessively oversized channels may even have a negative influence. For example, the shuffled Swin-Ti (28.8M, 4.3G MACs) only reaches a slightly higher accuracy at 81.4\% than its vanilla version (28.8M, 4.5G MACs) at 80.8\%. We argue that this module is most beneficial for tiny models that lack feature representations. As the feature representations get more complicated when the model scales up, the influence of this module vanishes.

\subsection{Ablation study}
We propose two crucial components in the channel shuffle module: the independent channel shuffle process and the channel re-scaling. In this section, we conduct ablation studies to evaluate the impact of these two design choices. 
\begin{table}[t]
\begin{center}
\caption{Ablation study on the two components} 
\setlength{\tabcolsep}{3.5pt}
\label{tab:ablationstudy}
\begin{tabular}{|l|c|c|c|c|c|}
  \hline
  Method & \makecell{channel \\ shuffle} & \makecell{channel \\ re-scale} & \#Params & \#MACs & \makecell{Top-1 \\ Acc.}\\
  \hline
  \multirow{3}*{T2T-ViT-7} & & & 4.3M & 1.1G & 71.7\% \\
  ~ & $\surd$ & & 4.7M & 1.3G & 72.5\% \\
  ~ & $\surd$ & $\surd$ & 4.7M & 1.3G & 74.4\% \\
  \hline
  \multirow{3}*{DeiT-Tiny} & & & 5.7M & 1.3G & 72.2\% \\
  ~ & $\surd$ & & 6.1M & 1.3G & 72.9\% \\
  ~ & $\surd$ & $\surd$ & 6.1M & 1.3G & 74.4\% \\
  \hline
  \multirow{3}*{Swin-ExtraTiny} & & & 6.9M & 1.1G & 74.8\% \\
  ~ & $\surd$ & & 7.2M & 1.0G & 75.8\% \\
  ~ & $\surd$ & $\surd$ & 7.2M & 1.0G & 77.8\% \\
  \hline
\end{tabular}
\end{center}
\end{table}

\noindent \textbf{Channel shuffle} Table \ref{tab:ablationstudy} demonstrates that without re-scaling, simply integrating the channel shuffle module into the vision Transformer leads to minor performance improvements. This is attributed to the fact that the channel shuffle process can introduce more imbalanced features when the number of feature channels is small. Despite marginal performance enhancement, merely adopting the channel shuffle module still helps to reach higher accuracy.

\noindent \textbf{Channel re-scaling} Table \ref{tab:ablationstudy} highlights the importance of channel re-scaling in this module, as it improves the outcome of the channel shuffle. For example, the shuffled Swin-ExtraTiny model with channel re-scaling surpasses both the original version and the non-scaling version, achieving a 3.0\% and 2.0\% increase in top-1 accuracy, respectively.

\section{Conclusion}
This paper presents an efficient module designed to enhance tiny ViT models. The proposed channel shuffle module expands the number of feature channels of a tiny ViT to improve its feature representation ability. In each layer, the feature channels are partitioned into two groups called the \textit{Attended} group and the \textit{Idle} group. Only the \textit{Attended} group participates in each layer's calculation while the \textit{Idle} group maintains the same until the end of the layer. The channel shuffle module effectively leverages channel shuffle operation to exchange information between the two groups, which contributes to enriched channel-wise information without introducing significant additional computational complexity. Experimental results demonstrate the effectiveness and generalizability of our module in improving both plain and hierarchical tiny ViTs.

\bibliographystyle{IEEEtran}
\bibliography{main}

\end{document}